\documentclass[runningheads]{llncs}
\usepackage{graphicx}
\usepackage{tabularx} 
\usepackage{subcaption}
\usepackage{caption}  
\usepackage{booktabs}
\usepackage{capt-of}
\usepackage{amssymb}

\newcommand{\matr}[1]{\mathbf{#1}}
\begin{document}

\section*{Acknowledgement}
This article was first published in Greenspan, H., et al. Medical Image Computing and Computer Assisted Intervention – MICCAI 2023. MICCAI 2023. Lecture Notes in Computer Science, vol 14224. Springer, Cham. The final authenticated version is available online at: \url{https://doi.org/10.1007/978-3-031-43904-9\_16}.
\title{Detecting domain shift in multiple instance learning for digital pathology using Fr\'echet Domain Distance\thanks{Supported by Swedish e-Science Research Center, VINNOVA, The CENIIT career development program at Linköping University, and Wallenberg AI, WASP funded by the Knut and Alice Wallenberg Foundation.}}
\titlerunning{Detecting domain shift in MIL}
%
\author{Milda Pocevičiūtė\inst{1,2}\orcidID{0000-0002-8734-6500} \and
Gabriel Eilertsen\inst{1,2}\orcidID{0000-0002-9217-9997} \and
Stina Garvin\inst{3} \and
Claes Lundström\inst{1,2,4}\orcidID{0000-0002-9368-0177}}
%
%
\authorrunning{M. Pocevičiūtė et al.}
%
\institute{Center for Medical Imaging Science and Visualization, Linköping University, University Hospital, Linköping, Sweden \and
Department of Science and Technology, Linköping University, Sweden
\email{\{milda.poceviciute, gabriel.eilertsen, claes.lundstrom\}@liu.se}\\ \and
Department of Clinical Pathology, and Department of Biomedical and Clinical Sciences, Linköping University, Linköping, Sweden\\
\email{garvin.stina@regionostergotland.se} \and
Sectra AB, Linköping, Sweden}
\maketitle              
\begin{abstract}
Multiple-instance learning (MIL) is an attractive approach for digital pathology applications as it reduces the costs related to data collection and labelling. However, it is not clear how sensitive MIL is to clinically realistic domain shifts, i.e., differences in data distribution that could negatively affect performance, and if already existing metrics for detecting domain shifts work well with these algorithms. We trained an attention-based MIL algorithm to classify whether a whole-slide image of a lymph node contains breast tumour metastases. The algorithm was evaluated on data from a hospital in a different country and various subsets of this data that correspond to different levels of domain shift. Our contributions include showing that MIL for digital pathology is affected by clinically realistic differences in data, evaluating which features from a MIL model are most suitable for detecting changes in performance, and proposing an unsupervised metric named Fréchet Domain Distance (FDD) for quantification of domain shifts. Shift measure performance was evaluated through the mean Pearson correlation to change in classification performance, where FDD achieved 0.70 on 10-fold cross-validation models. The baselines included Deep ensemble, Difference of Confidence, and Representation shift which resulted in 0.45, -0.29, and 0.56 mean Pearson correlation, respectively. FDD could be a valuable tool for care providers and vendors who need to verify if a MIL system is likely to perform reliably when implemented at a new site, without requiring any additional annotations from pathologists.

\end{abstract}
\section{Introduction}
The spreading digitalisation of pathology labs has enabled the development of deep learning (DL) tools that can assist pathologists in their daily tasks. However, supervised DL methods require detailed annotations in whole-slide images (WSIs) which is time-consuming, expensive and prone to inter-observer disagreements \cite{Gomes2014}. Multiple instance learning (MIL) alleviates the need for detailed annotations and has seen increased adoption in recent years. MIL approaches have proven to work well in academic research on histopathology data \cite{campanella2019,CLAM,SU2022} as well as in commercial applications \cite{Silva2021}. Most MIL methods for digital pathology employ an attention mechanism as it increases the reliability of the algorithms, which is essential for successful clinical adoption \cite{javed2022additive}.

Domain shift in DL occurs when the data distributions of testing and training differs \cite{MorenoTorres2012,Zhou2022}. This remains a significant obstacle to the deployment of DL applications in clinical practice \cite{Guan2021}. To address this problem previous work either use domain adaptation when data from the target domain is available \cite{Xiaofeng2022}, or domain generalisation when the target data is unavailable \cite{Zhou2022}. Domain adaptation has been explored in the MIL setting too \cite{Prabono2022,Praveen2020,Song2020}. However, it may not be feasible to perform an explicit domain adaptation, and an already adapted model could still experience problems with domain shifts. Hence, it is important to provide indications of the expected performance on a target dataset without requiring annotations \cite{Elsahar2019,Schelter2020b}. Another related topic is out-of-distribution (OOD) detection \cite{yang2021generalized} which aims to detect individual samples that are OOD, in contrast to our objective of estimating a difference of expected performances between some datasets.
For supervised algorithms, techniques of uncertainty estimation have been used to measure the effect of domain shift \cite{Elder2020,Lakshminarayanan2016,Maggio2022} and to improve the robustness of predictions \cite{Martinez_2019_CVPR_Workshops,poceviciute2022,Thagaard2020}. However, the reliability of uncertainty estimates can also be negatively affected by domain shifts \cite{Hoebel2022,Tomani_2021_CVPR}. Alternatively, a drop in performance can be estimated by comparing the model's softmax outputs \cite{Guillory2021} or some hidden features \cite{Rabanser2019,Stacke2021} acquired on in-domain and domain shift datasets. Although such methods have been demonstrated for supervised algorithms, as far as we know no previous work has explored domain shift in the specific context of MIL algorithms. Hence, it is not clear how well they will work in such a scenario.

In this work, we evaluate an attention-based MIL model on unseen data from a new hospital and propose a way to quantify the domain shift severity. The model is trained to perform binary classification of WSIs from lymph nodes of breast cancer patients. We split the data from the new hospital into several subsets to investigate clinically realistic scenarios triggering different levels of domain shift. We show that our proposed unsupervised metric for quantifying domain shift correlates best with the changes in performance, in comparison to multiple baselines. The approach of validating a MIL algorithm in a new site without collecting new labels can greatly reduce the cost and time of quality assurance efforts and ensure that the models perform as expected in a variety of settings. The novel contributions of our work can be summarised as:
\begin{enumerate}
    \item Proposing an unsupervised metric named Fr\'echet Domain Distance (FDD) for quantifying the effects of domain shift in attention-based MIL;
    \item Showing how FDD can help to identify subsets of patient cases for which MIL performance is worse than reported on the in-domain test data;
    \item Comparing the effectiveness of using uncertainty estimation versus learnt representations for domain shift detection in MIL.
\end{enumerate}

\section{Methods}

Our experiments center on an MIL algorithm with attention developed for classification in digital pathology. The two main components of our domain shift quantification approach are the selection of MIL model features to include and the similarity metric to use, described below.

\subsection{MIL method}
As the MIL method for our investigation, we chose the clustering-constrained-attention MIL (CLAM)~\cite{CLAM} because it 
well represents an architecture of MIL with attention, meaning that our approach can equally be applied to many other such methods.

\subsection{MIL features} \label{sec_feat_selection}
We explored several different feature sets that can be extracted from the attention-based MIL framework: learnt embedding of the instances (referred to as \textit{patch features}), and penultimate layer features (\textit{penultimate features}). 
A study is conducted to determine the best choices for type and amount of patch features. As a baseline, we take a mean over all patches ignoring their attention scores (\textit{mean patch features}). Alternatively, the patch features can be selected based on the attention score assigned to them. \textit{Positive evidence} or \textit{Negative evidence} are defined as the $K$ patch features that have the $K$ highest or lowest attention scores, respectively. \textit{Combined evidence} is a combination of an equal number of patch features with the highest and lowest attention scores. To test if the reduction of the number of features in itself has a positive effect on domain shift quantification, we also compare with $K$ randomly selected patch features. 

\subsection{Fr\'echet Domain Distance}
Fr\'echet Inception Distance (FID) \cite{FID_paper} is commonly used to measure similarity between real and synthetically generated data. Inspired by FID, we propose a metric named Fr\'echet Domain Distance (FDD) for evaluating if a model is experiencing a drop in performance on some new dataset. The Fr\'echet distance (FD) between two multivariate Gaussian variables with means $\mu_1$, $\mu_2$ and covariance matrices $\matr{C}_1$, $\matr{C}_2$  is defined as \cite{frechet_distance}:
\begin{equation}
\label{eq:fd}
FD((\mu_{1},\matr{C}_{1}),(\mu_{2},\matr{C}_{2}))=\|\mu_{1} - \mu_{2}\|^{2} + Tr(\matr{C}_{1} +\matr{C}_{2} - 2 (\matr{C}_{1} \matr{C}_{2})^{\frac{1}{2}}).
\end{equation}
We are interested in using the FD for measuring the domain shift between different WSI datasets $\mathcal{X}_d$. To this end, we extract features from the MIL model applied to all the WSIs in $\mathcal{X}_d$, and arrange these in a feature matrix $\matr{M}_d \in \mathbb{R}^{W_d \times J}$, where $W_d$ is the number of WSIs in $\mathcal{X}_d$ and $J$ is the number of features extracted by the MIL model for one WSI $x \in \mathcal{X}_d$.
For penultimate layer and mean patch features, we can apply this strategy directly. However, for positive, negative, and combined evidence we have $K$ feature matrices $\matr{M}_{d,k}$, with $K$ feature descriptions for each WSI.
To aggregate evidence features over a WSI, we take the mean $\matr{M}_d^{K} = \frac{1}{K} \sum_{k=1}^K \matr{M}_{d,k}$.
Now, we can compute means $\mu_{d}^{K}$ and covariance matrices $\matr{C}_{d}^{K}$ from $\matr{M}_d^{K}$ extracted from two datasets $d = 1$ and $d = 2$. By measuring the FD, we arrive at our proposed FDD: 
\begin{equation}
\label{eq:fdd}
FDD_{K}(\mathcal{X}_1,\mathcal{X}_2) = FD((\mu_{1}^{K},\matr{C}_{1}^{K}),(\mu_{2}^{K},\matr{C}_{2}^{K})).
\end{equation}
$FDD_K$ uses the $K$ aggregated positive evidence features, but in the results we also compare to $\matr{M}_d$ described from penultimate features, mean patch features, and the other evidence feature selection strategies.

\section{Datasets}\label{sec_data}
Grand Challenge Camelyon data \cite{camelyon17} is used for model training (770 WSIs of which 293 WSIs contain metastases) and in-domain testing (629 WSIs of which 289 WSIs contain metastases). For domain shift test data, we extracted 302 WSIs (of which 111 contain metastases) from AIDA BRLN dataset~\cite{jarkman019}. To evaluate clinically realistic domain shifts, we divided the dataset in two different ways, creating four subsets:
\begin{enumerate}
    \item[1a.] 161 WSIs from sentinel node biopsy cases (54 WSIs with metastases): a small shift as it is the same type of lymph nodes as in Grand Challenge Camelyon data \cite{camelyon17}.
    \item[1b.] 141 WSIs from axillary nodes dissection cases (57 WSIs with metastases): potentially large shift as some patients have already started neoadjuvant treatment as well as the tissue may be affected from the procedure of sentinel lymph node removal.
    \item[2a.] 207 WSIs with ductal carcinoma (83 WSIs with metastases): a small shift as it is the most common type of carcinoma and relatively easy to diagnose.
    \item[2b.] 68 WSIs with lobular carcinoma (28 WSIs with metastases): potentially large shift as it is a rare type of carcinoma and relatively difficult to diagnose.
\end{enumerate}

The datasets of lobular and ductal carcinomas each contain 50 \% of WSIs from sentinel and axillary lymph node procedures. The sentinel/axillary division is motivated by the differing DL prediction performance on such subsets, as observed by Jarkman et al.~\cite{Jarkman2022}. Moreover, discussions with pathologists led to the conclusion that it is clinically relevant to evaluate the performance difference between ductal and lobular carcinoma. Our method is intended to avoid requiring dedicated WSI labelling efforts. We deem that the information needed to do this type of subset divisions would be available without labelling since the patient cases in a clinical setting would already contain such information. All datasets are publicly available to be used in legal and ethical medical diagnostics research.

\section{Experiments}
The goal of the study is to evaluate how well $FDD_{K}$ and the baseline methods correlate with the drop in classification performance of attention-based MIL caused by several potential sources of domain shifts. In this section, we describe the experiments we conducted.

\subsection{MIL training} \label{subsec_mil_training}
We trained, with default settings, 10 CLAM models to classify WSIs of breast cancer metastases using a 10-fold cross-validation (CV) on the training data. The test data was kept the same for all 10 models. 
The classification performance is evaluated using the area under receiver operating characteristic curve (ROC-AUC) and Matthews correlation coefficient (MCC) \cite{MCC_benefits}. Following the conclusions of \cite{MCC_benefits} that MCC well represents the full confusion matrix and the fact that in clinical practice a threshold needs to be set for a classification decision, MCC is used as a primary metric of performance for domain shift analysis while ROC-AUC is reported for completeness. Whereas extremely large variations in label prevalence could reduce the reliability of the MCC metric, this is not the case here as label prevalence is similar (35-45\%) in our test datasets. For Deep ensemble \cite{Lakshminarayanan2016} we trained 4 additional CLAM models for each of the 10 CV folds.

\subsection{Domain shift quantification}
As there is no related work on domain shift detection in the MIL setting, we selected methods developed for supervised algorithms as baselines:
\begin{itemize}
    \item The model's accumulated uncertainty between two datasets. Deep ensemble \cite{Lakshminarayanan2016} (DE) and Difference in Confidence with entropy \cite{Guillory2021} (DoC) compare the mean entropy over all data points. DE uses an ensemble to estimate better-calibrated uncertainty than the single model in DoC.
    \item The accumulated confidence of a model across two datasets. DoC \cite{Guillory2021} can be measured on the mean softmax scores of two datasets. A large difference indicates a potential drop in performance. 
    \item The hidden features produced by an algorithm. Representation Shift \cite{Stacke2021} (RS) has shown promising results in detecting domain shift in convolutional neural networks and it is the method most similar to FDD. However, it is not trivial which hidden features of MIL that are most suitable for this task, and we evaluate several options (see Section \ref{sec_feat_selection}) with both methods.
\end{itemize}

For all possible pairs of Camelyon and the other test datasets, and for the 10 CV models, we compute the domain shift measures and compare them to the observed drop in performance. The effectiveness is evaluated by Pearson correlation and visual investigation of corresponding scatter plots. All results are reported as mean and standard deviation over the 10-fold CV.

\section{Results}
\begin{figure}
\parbox[t]{0.45\linewidth}{\null
\centering
  \vskip-\abovecaptionskip
  \captionof{table}[t]{Classification performance reported in mean (standard deviation) of MCC and ROC-AUC metrics, computed over the 10-fold CV models. A threshold for MCC is determined on validation data.}%
  \label{tab_performance}
  \vskip\abovecaptionskip
  \begin{tabular}{|p{0.12\textwidth}|p{0.14\textwidth}|p{0.14\textwidth}|}
    \hline
    \textit{Dataset} & \textit{MCC} & \textit{ROC-AUC}  \\
    \hline
    Camelyon & 0.67 (0.02) & 0.87 (0.01) \\
    BRLN  &  0.61 (0.06) & 0.89 (0.01) \\     
    Sentinel &  0.67 (0.05) & 0.90 (0.01)  \\   
    Axillary & 0.56 (0.08) & 0.86 (0.01)  \\  
    Ductal & 0.72 (0.04) & 0.92 (0.01) \\    
    Lobular &  0.61 (0.04) & 0.85 (0.03) \\  
    \hline
  \end{tabular}
}
\parbox[t]{0.06\linewidth}{\null \hfill}
\parbox[t]{0.47\linewidth}{\null
\hfill\includegraphics[width=0.47\textwidth]{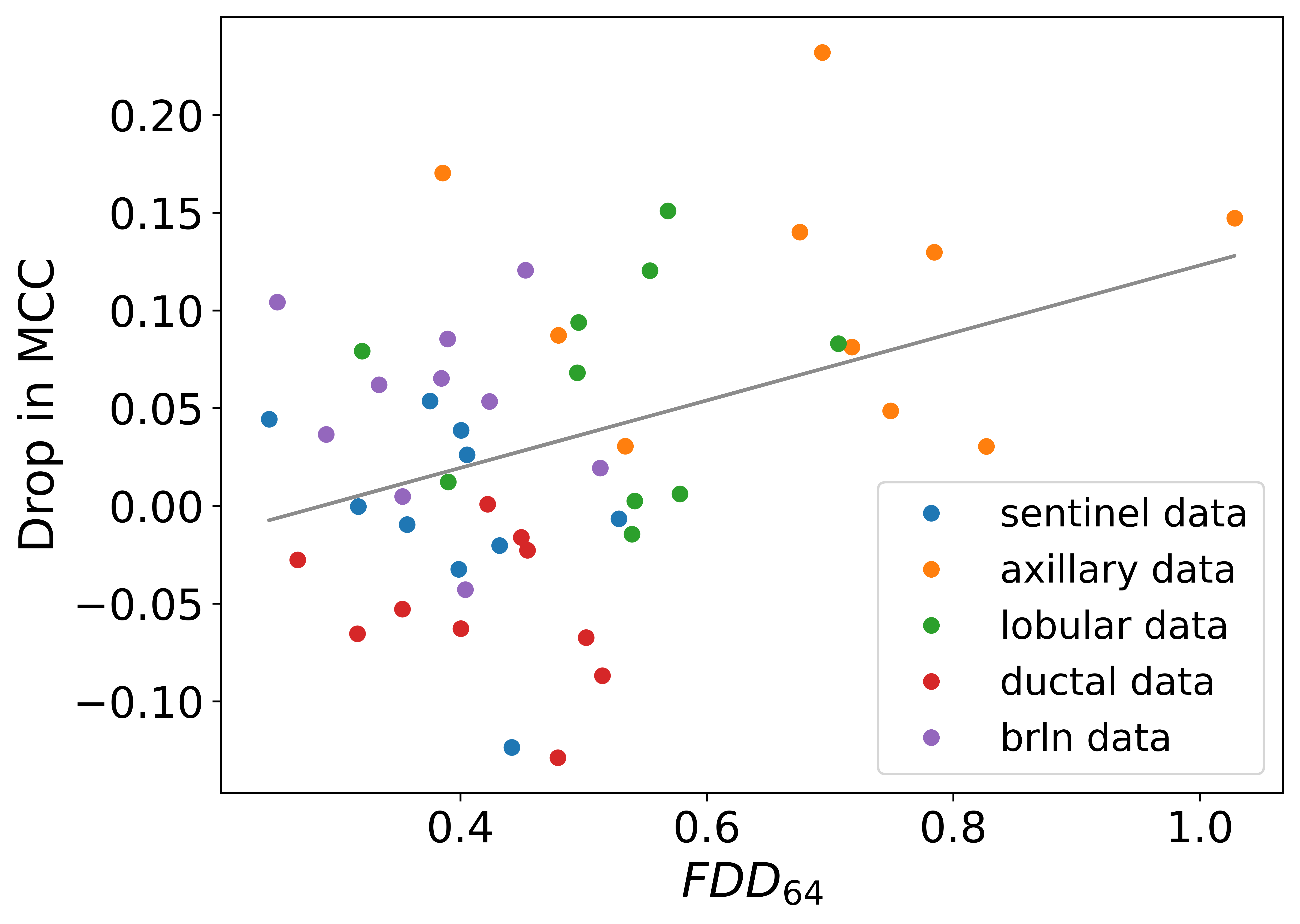}
  \captionof{figure}{Scatter plot of $FDD_{64}$ against drop in MCC for all model-dataset combinations. A fitted line is included for better interpretability.}%
  \label{fig_scatter}
}
\end{figure}

The first part of our results is the performance of the WSI classification task across the subsets, summarized in Table \ref{tab_performance}. While showing similar trends, there is some discrepancy in the level of domain shift represented by the datasets due to the differences between the MCC and ROC-AUC measures. 

As we deemed MCC to better represent the clinical use situation (see Section \ref{subsec_mil_training}), it was used for our further evaluations. 
Overall, the performance is in line with previously published work \cite{CLAM,SU2022}.

We observe the largest domain shift in terms of MCC on axillary nodes followed by lobular carcinoma and full BRLN datasets. There seems to be no negative effect from processing the sentinel nodes data. CLAM models achieved better performance on ductal carcinoma compared to the in-domain Camelyon test data. 

Table \ref{tab_pearson_corrs} summarises the Pearson correlation between the change in performance, i.e., the MCC difference between Camelyon and other test datasets, and the domain shift measures for the same pairs. $FDD_{64}$ outperforms the baselines substantially, and has the smallest standard deviation. Fig.~\ref{fig_scatter} shows how individual drop in performance of model-dataset combinations are related to the $FDD_{64}$ metric. For most models detecting larger drop in performance ($>0.05$) is easier on axillary lymph nodes data than on any other analysed dataset. 

\begin{table}
\centering
\caption{Pearson correlations between domain shift measure and difference in performance (MCC metric) of Camelyon test set and the other datasets. The mean and standard deviation values are computed over the 10 CV folds.}\label{tab_pearson_corrs}
\begin{tabular}{|l|l|l|}
\hline
\textit{Measure} & \textit{Features used} &  \textit{Pearson correlation}  \\
\hline
DoC & Softmax score & -0.29 (0.46) \\
DoC & Entropy score & -0.32 (0.47) \\
DE & Entropy score & 0.45 (0.38) \\
RS & Penultimate feat. & 0.48 (0.25) \\
RS & Mean patch feat. & 0.41 (0.32) \\
$RS_{64}$ & Positive evidence & 0.56 (0.19) \\
FD & Mean patch feat. & 0.49 (0.28) \\
FD & Penultimate feat. & 0.61 (0.19) \\
$FDD_{64}$ (our) & Positive evidence  & 0.70 (0.15) \\
\hline
\end{tabular}
\end{table}

\begin{figure}[t]
\centering
\includegraphics[width=\textwidth]{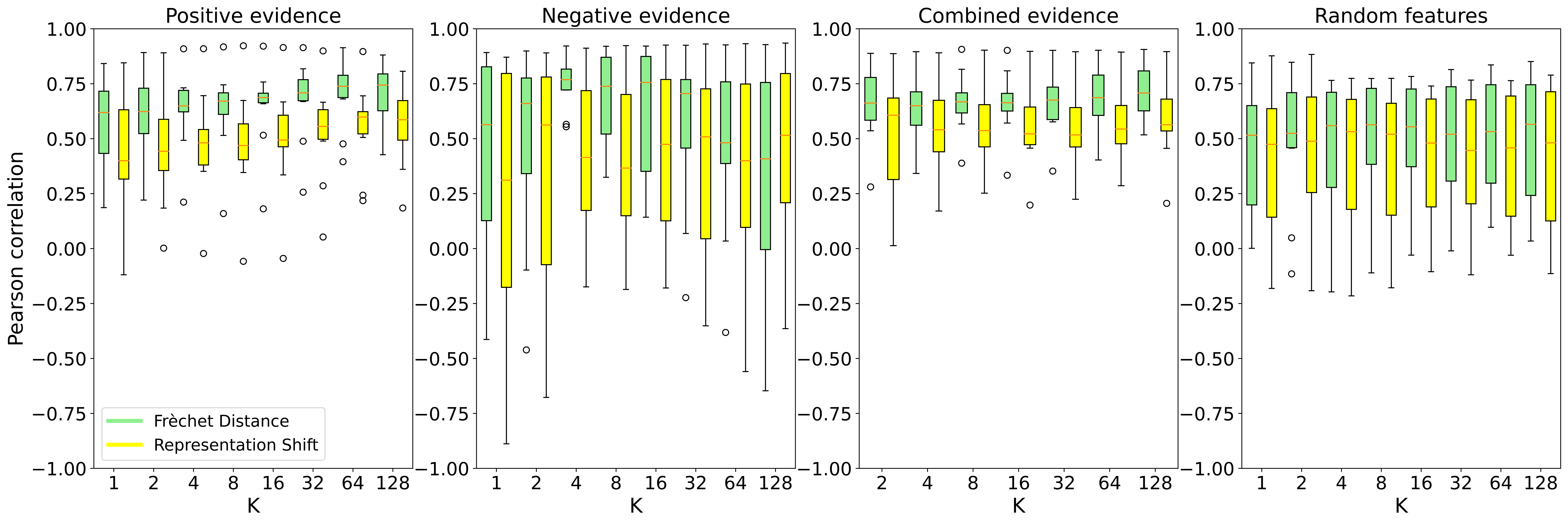}
\caption{Box plots of Pearson correlations achieved by Fr\'echet distance and Representation shift metric using attention-based features, i.e., positive, negative, combined evidence, and randomly selected features. Varying number of extracted patch representations $K$ is considered: from 1 to 128. The reported results are over the 10 cross validation models.} \label{fig_k_features_h}
\end{figure}

A study of the number and type of MIL attention-based features and FD and RS metrics is presented in Fig.~\ref{fig_k_features_h}. The baseline of randomly selecting patch features resulted in the worst outcome on domain shift detection. Negative evidence with FD achieved high Pearson correlation when $K=4$. However, the results were among the worst with any other number of $K$. Both combined and positive evidence achieved peak performance  of 0.68 (0.17) and 0.70 (0.13), respectively, when FD and $K=64$ were used. We conclude that in our setup the best and most reliable performance of domain shift quantification is achieved by positive evidence with FD and $K=64$, i.e. $FDD_{64}$. 

\section{Discussion and Conclusion} \label{sec_discission}

\textbf{MIL is affected by domain shift}
Some previous work claim that MIL is more robust to domain shift as it is trained on more data due to the reduced costs of data annotation \cite{campanella2019,CLAM}. We argue that domain shift will still be a factor to consider as an algorithm deployed in clinical practice is likely to encounter unseen varieties of data. However, it may require more effort to determine what type of changes in data distribution are critical. Our results show that domain shift is present between the WSIs from the same hospital (Camelyon data) and another medical centre (BRLN data). However, as clinically relevant subsets of BRLN data are analysed, stark differences in performance and reliability (indicated by the standard deviation) are revealed. Therefore, having a reliable metric for unsupervised domain shift quantification could bring value for evaluating an algorithm at a new site. \newline

\noindent\textbf{Domain shift detection for supervised DL struggles for MIL}
An important question is whether can we apply existing techniques from supervised learning algorithms in the MIL setting. The evaluated baselines that use uncertainty and confidence aggregation for domain shift detection, i.e., DE and DoC, showed poor ability to estimate the experienced drop in performance (see table \ref{tab_pearson_corrs}). It is known that supervised DL often suffers from overconfident predictions \cite{Guo2018}. This could be a potential cause for the observed poor results by DE and DoC in our experiments. Further investigation on how to improve calibration could help to boost the applicability of uncertainty and confidence measures.\newline

\noindent\textbf{The proposed $FDD_K$ measure outperforms alternatives}
The highest Pearson correlation between change in performance and a distance metric is achieved by Fr\'echet distance with 64 positive evidence features, $FDD_{64}$ (see Table \ref{tab_pearson_corrs}). $RS_{64}$ approach performed better than uncertainty/confidence-based methods but still was substantially worse than $FDD_{64}$. Furthermore, $FDD_{64}$ resulted in the smallest standard deviation which is an important indicator of the reliability of the metric. Interestingly, using penultimate layer features, which combine all patch features and attention scores, resulted in much worse outcome than $FDD_{64}$, 0.61 versus 0.70. Thus, it seems a critical component in domain shift measurement in attention-based MIL is to correctly make use of the attention scores. 
From Fig. \ref{fig_scatter} we can see that if we further investigated all model-dataset combinations that resulted in $FDD_{64}$ above 0.5, we would detect many cases with a drop in performance larger than 0.05. However, the drop is easier to detect on axillary and lobular datasets compared to others.
An interesting aspect is that the performance was better for the out-of-domain ductal subset compared to in-domain Camelyon WSIs. In practical applications, it may be a problem when the domain shift quantification cannot separate between shifts having positive or negative effect on performance. Such differentiation could be the topic of future work. \newline

\noindent\textbf{Conclusion} We carried out a study on how clinically realistic domain shifts affect attention-based MIL for digital pathology. The results show that domain shift may raise challenges in MIL algorithms. Furthermore, there is a clear benefit of using attention for feature selection and our proposed $FDD_{K}$ metric for quantification of expected performance drop. Hence, $FDD_{K}$ could aid care providers and vendors in ensuring safe deployment and operation of attention-based MIL in pathology laboratories.

\newpage
\bibliographystyle{splncs04}
\bibliography{mybibliography}

\begin{thebibliography}{10}
\providecommand{\url}[1]{\texttt{#1}}
\providecommand{\urlprefix}{URL }
\providecommand{\doi}[1]{https://doi.org/#1}

\bibitem{campanella2019}
Campanella, G., et~al.: Clinical-grade computational pathology using weakly supervised deep learning on whole slide images. Nature Medicine  \textbf{25}(8),  1301--1309 (2019)

\bibitem{MCC_benefits}
Chicco, D., Jurman, G.: The advantages of the matthews correlation coefficient (mcc) over f1 score and accuracy in binary classification evaluation. BMC Genomics  \textbf{21}(1),  1 -- 13 (2020)

\bibitem{frechet_distance}
Dowson, D., Landau, B.: The fréchet distance between multivariate normal distributions. Journal of Multivariate Analysis  \textbf{12}(3),  450--455 -- 455 (1982)

\bibitem{Elder2020}
Elder, B., Arnold, M., Murthi, A., Navratil, J.: Learning prediction intervals for model performance. In: Proceedings of the AAAI Conference on Artificial Intelligence (2020)

\bibitem{Elsahar2019}
Elsahar, H., Gall{\'e}, M.: To annotate or not? predicting performance drop under domain shift. In: Proceedings of the 2019 Conference on Empirical Methods in Natural Language Processing and the 9th International Joint Conference on Natural Language Processing (EMNLP-IJCNLP). pp. 2163--2173 (2019)

\bibitem{Gomes2014}
Gomes, D.S., Porto, S.S., Balabram, D., Gobbi, H.: Inter-observer variability between general pathologists and a specialist in breast pathology in the diagnosis of lobular neoplasia, columnar cell lesions, atypical ductal hyperplasia and ductal carcinoma in situ of the breast. Diagnostic Pathology  \textbf{9}(1), ~121 (2014)

\bibitem{Guan2021}
Guan, H., Liu, M.: Domain adaptation for medical image analysis: a survey. IEEE Transactions on Biomedical Engineering  \textbf{69}(3),  1173--1185 (2021)

\bibitem{Guillory2021}
Guillory, D., Shankar, V., Ebrahimi, S., Darrell, T., Schmidt, L.: Predicting with confidence on unseen distributions. In: 2021 IEEE/CVF International Conference on Computer Vision (ICCV). pp. 1114--1124 (2021)

\bibitem{Guo2018}
Guo, C., Pleiss, G., Sun, Y., Weinberger, K.Q.: On calibration of modern neural networks. In: Proceedings of the 34th International Conference on Machine Learning. Proceedings of Machine Learning Research, vol.~70, pp. 1321--1330. PMLR (2017)

\bibitem{FID_paper}
Heusel, M., Ramsauer, H., Unterthiner, T., Nessler, B., Hochreiter, S.: Gans trained by a two time-scale update rule converge to a local nash equilibrium  (2017)

\bibitem{Hoebel2022}
Hoebel, K., et~al.: Do i know this? segmentation uncertainty under domain shift. In: Proceedings of SPIE, the International Society for Optical Engineering. vol. 12032, pp. 1203211 -- 1203211--16 (2022)

\bibitem{jarkman019}
Jarkman, S., Lindvall, M., Hedlund, J., Treanor, D., Lundstr{\"o}m, C., van~der Laak, J.: Axillary lymph nodes in breast cancer cases. (2019)

\bibitem{Jarkman2022}
Jarkman, S., et~al.: Generalization of deep learning in digital pathology: Experience in breast cancer metastasis detection. Cancers  \textbf{14}(21), ~5424 (2022)

\bibitem{javed2022additive}
Javed, S.A., Juyal, D., Padigela, H., Taylor-Weiner, A., Yu, L., aaditya prakash: Additive {MIL}: Intrinsically interpretable multiple instance learning for pathology. In: Advances in Neural Information Processing Systems (2022)

\bibitem{Lakshminarayanan2016}
Lakshminarayanan, B., Pritzel, A., Blundell, C.: Simple and scalable predictive uncertainty estimation using deep ensembles. In: Advances in Neural Information Processing Systems. vol.~30. Curran Associates, Inc. (2017)

\bibitem{camelyon17}
Litjens, G., et~al.: {1399 H\&E-stained sentinel lymph node sections of breast cancer patients: the CAMELYON dataset}. GigaScience  \textbf{7}(6) (2018)

\bibitem{CLAM}
Lu, M.Y., et~al.: Data-efficient and weakly supervised computational pathology on whole-slide images. Nature Biomedical Engineering  \textbf{5}(6),  555 -- 570 (2021)

\bibitem{Maggio2022}
Maggio, S., Bouvier, V., Dreyfus-Schmidt, L.: Performance prediction under dataset shift. In: 2022 26th International Conference on Pattern Recognition (ICPR). pp. 2466--2474 (2022)

\bibitem{Martinez_2019_CVPR_Workshops}
Martinez, C., et~al.: Segmentation certainty through uncertainty: Uncertainty-refined binary volumetric segmentation under multifactor domain shift. In: Proceedings of the IEEE/CVF Conference on Computer Vision and Pattern Recognition (CVPR) Workshops (2019)

\bibitem{MorenoTorres2012}
Moreno-Torres, J.G., et~al.: A unifying view on dataset shift in classification. Pattern Recognition  \textbf{45}(1),  521--530 (2012)

\bibitem{poceviciute2022}
Pocevičiūtė, M., Eilertsen, G., Jarkman, S., Lundström, C.: Generalisation effects of predictive uncertainty estimation in deep learning for digital pathology. Scientific Reports  \textbf{12}(1),  1 -- 15 (2022)

\bibitem{Prabono2022}
Prabono, A.G., Yahya, B.N., Lee, S.L.: Multiple-instance domain adaptation for cost-effective sensor-based human activity recognition. Future Generation Computer Systems  \textbf{133},  114--123 (2022)

\bibitem{Praveen2020}
Praveen, R.G., Granger, E., Cardinal, P.: Deep weakly supervised domain adaptation for pain localization in videos. In: 2020 15th IEEE International Conference on Automatic Face and Gesture Recognition (FG 2020). pp. 473--480. IEEE (2020)

\bibitem{Rabanser2019}
Rabanser, S., Günnemann, S., Lipton, Z.: Failing loudly: An empirical study of methods for detecting dataset shift. In: NeurIPS 2019 (2019)

\bibitem{Schelter2020b}
Schelter, S., Rukat, T., Bie{\ss}mann, F.: Learning to validate the predictions of black box classifiers on unseen data. In: Proceedings of the 2020 ACM SIGMOD International Conference on Management of Data. pp. 1289--1299 (2020)

\bibitem{Silva2021}
Silva, L.M., et~al.: Independent real‐world application of a clinical‐grade automated prostate cancer detection system. In: Journal of pathology. vol.~254, pp. 147 -- 158 (2021)

\bibitem{Song2020}
Song, R., Cao, P., Yang, J., Zhao, D., Zaiane, O.R.: A domain adaptation multi-instance learning for diabetic retinopathy grading on retinal images. In: 2020 IEEE international conference on bioinformatics and biomedicine (BIBM). pp. 743--750. IEEE (2020)

\bibitem{Stacke2021}
Stacke, K., Eilertsen, G., Unger, J., Lundstrom, C.: Measuring domain shift for deep learning in histopathology. In: IEEE journal of biomedical and health informatics. p.~325. No.~2 (2021)

\bibitem{SU2022}
Su, Z., et~al.: Attention2majority: Weak multiple instance learning for regenerative kidney grading on whole slide images. Medical Image Analysis  \textbf{79},  102462 (2022)

\bibitem{Thagaard2020}
Thagaard, J., Hauberg, S., van~der Vegt, B., Ebstrup, T., Hansen, J.D., Dahl, A.B.: Can you trust predictive uncertainty under real dataset shifts in digital pathology?. In: Lecture notes in computer science. pp. 824--833 (2020)

\bibitem{Tomani_2021_CVPR}
Tomani, C., Gruber, S., Erdem, M.E., Cremers, D., Buettner, F.: Post-hoc uncertainty calibration for domain drift scenarios. In: Proceedings of the IEEE/CVF Conference on Computer Vision and Pattern Recognition (CVPR). pp. 10124--10132 (2021)

\bibitem{Xiaofeng2022}
Xiaofeng, L., et~al.: Deep unsupervised domain adaptation: A review of recent advances and perspectives. APSIPA Transactions on Signal and Information Processing  \textbf{11}(1) (2022)

\bibitem{yang2021generalized}
Yang, J., Zhou, K., Li, Y., Liu, Z.: Generalized out-of-distribution detection: A survey. arXiv preprint arXiv:2110.11334  (2021)

\bibitem{Zhou2022}
Zhou, K., et~al.: Domain generalization: A survey. IEEE Transactions on Pattern Analysis and Machine Intelligence  (2022)

\end{thebibliography}
\newpage

\begin{center}
\Huge \textbf{Supplementary material}
\end{center}

\begin{figure}
\centering
\includegraphics[width=\textwidth]{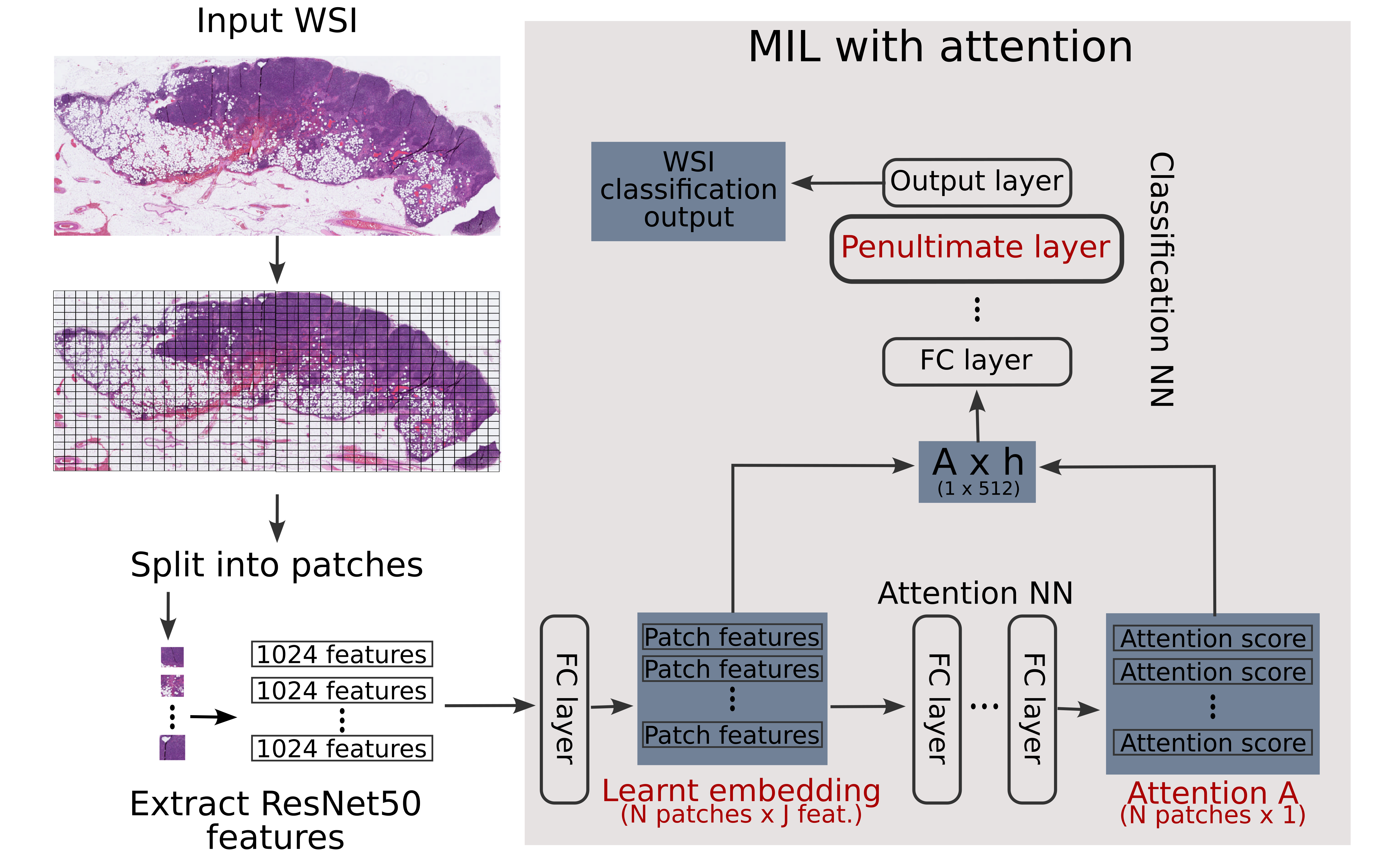}
\caption{Visualisation of an attention MIL framework for digital pathology. The features that are used in our experiments are marked in red. FC stands for Fully Connected.} \label{fig_clam_features}
\end{figure}

\begin{table}
\centering
\caption{CLAM parameters and their values used in the experiments}\label{tab_clam_params}
\begin{tabular}{|l|l|l|}
\hline
\textit{Parameter} & \textit{Parameter description} &  \textit{Value}  \\
\hline
max\_epochs & Maximum training epochs & 200 \\
lr & Learning rate & 0.0002 \\
reg & Learning rate decay & 0.00001 \\
bag\_loss & Classification loss function & cross entropy \\
model\_type & Single/multi branch CLAM & clam w. single branch \\
model\_size & Small or large CLAM model & small \\
use\_drop\_out & Enable dropout & True \\
weighted\_sample & Penultimate feat. & True \\
opt & Optimizer & ADAM \\
bag\_weight & Enables weighted sampling  & 0.7 \\
inst\_loss & Instance-level clustering loss function  & SVM \\
B & \# positive/negative patches to sample  & 32 \\
\hline
\end{tabular}
\end{table}

\begin{figure}
\centering
\includegraphics[width=\textwidth]{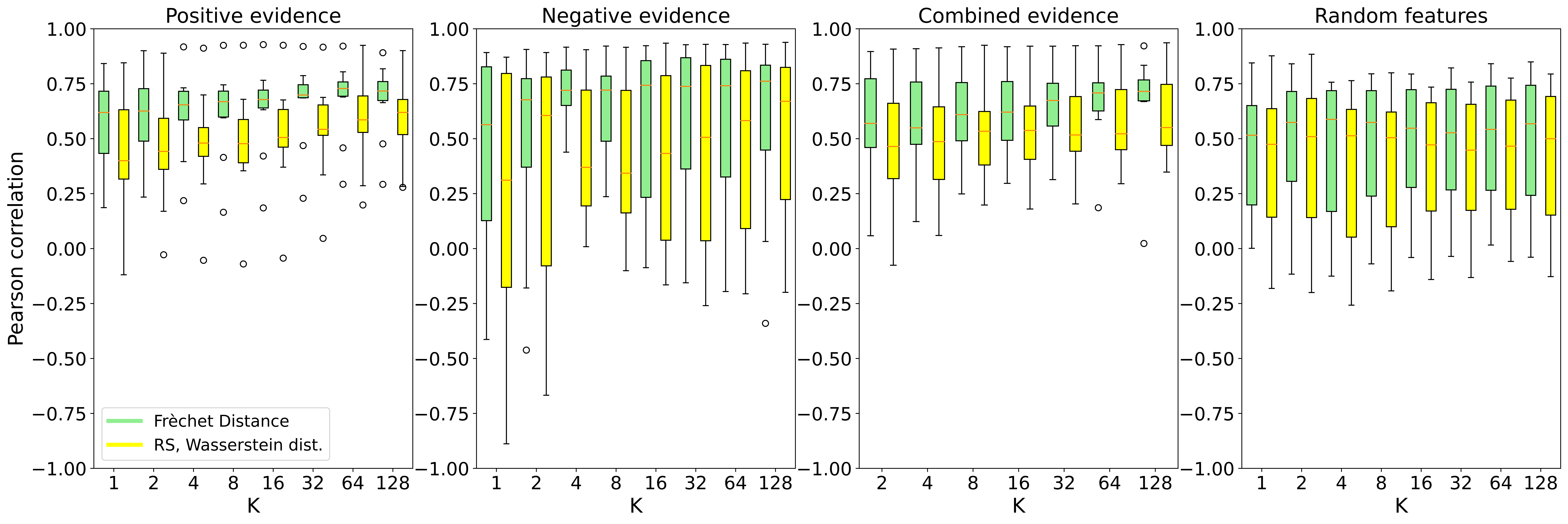}
\caption{Concatenated attention-based features: Pearson correlations achieved by Fr\'echet distance and Representation shift metric using positive, negative, combined evidence, and randomly selected features. Varying number of extracted patch representations $K$ is considered: from 1 to 128. The reported results are over the 10 CV models.} \label{fig_k_features_concat}
\end{figure}

\end{document}